\documentclass[11pt]{article}

\usepackage{fullpage}

\usepackage{amsfonts,epsfig,graphicx}
\usepackage{amsmath,amssymb,amsthm}
\usepackage{enumerate}

\usepackage{color}
\usepackage{macros}
\usepackage[bookmarks=true,colorlinks,citecolor=blue,urlcolor=blue]{hyperref}

\title{Splash: User-friendly Programming Interface for\\ Parallelizing Stochastic Algorithms}

\author{Yuchen Zhang ~~~~~~~~ Michael I. Jordan\\
Department of Electrical Engineering and Computer Science\\
University of California, Berkeley CA 94720\\
\texttt{\{yuczhang,jordan\}@eecs.berkeley.edu} 
}

\date{}

\begin{document}

\maketitle 

\begin{abstract}
Stochastic algorithms are efficient approaches to solving machine learning and optimization problems. In this paper, we propose a general framework called \emph{Splash} for parallelizing stochastic algorithms on multi-node distributed systems. Splash consists of a programming interface and an execution engine. Using the programming interface, the user develops sequential stochastic algorithms without concerning any detail about distributed computing. The algorithm is then automatically parallelized by a communication-efficient execution engine. We provide theoretical justifications on the optimal rate of convergence for parallelizing stochastic gradient descent. \splash\ is built on top of Apache Spark.
The real-data experiments on logistic regression, collaborative filtering and topic modeling verify that Splash yields order-of-magnitude speedup over single-thread stochastic algorithms and over state-of-the-art implementations on Spark.

\end{abstract}

\setlength{\belowdisplayskip}{5pt}
\setlength{\abovedisplayskip}{5pt}

\vspace{-5pt}
\section{Introduction}

Stochastic optimization algorithms process a large-scale dataset by sequentially processing random subsamples. This processing scheme makes the per-iteration cost of the algorithm much cheaper than that of batch processing algorithms while still yielding effective descent. Indeed, for convex optimization, the efficiency of stochastic gradient descent (SGD) and its variants has been established both in theory and in practice
\cite{zhang2004solving,bottou2010large,xiao2009dual,duchi2011adaptive,schmidt2013minimizing,johnson2013accelerating}.
For non-convex optimization, stochastic methods achieve state-of-the-art performance on a broad class of problems, including matrix factorization~\cite{koren2009matrix}, neural networks~\cite{krizhevsky2012imagenet} and representation learning~\cite{vincent2008extracting}. Stochastic algorithms are also widely used in the Bayesian setting for finding approximations to posterior distributions; examples include Markov chain Monte Carlo, expectation propagation~\cite{minka2001expectation} and stochastic variational inference~\cite{hoffman2013stochastic}. 

Although classical stochastic approximation procedures are sequential, it is clear that they also present opportunities for parallel and distributed implementations that may yield significant additional speedups.  One active line of research studies asynchronous parallel updating schemes in the setting of a lock-free shared memory~\cite{recht2011hogwild,duchi2012dual,liu2013asynchronous,
zhuang2013fast,ho2013more}. When the time delay of concurrent updates are bounded, it is known that such updates preserve statistical correctness~\cite{agarwal2011distributed,liu2013asynchronous}. Such asynchronous algorithms yield significant speedups on multi-core machines. On distributed systems connected by commodity networks, however, the communication requirements of such algorithms can be overly expensive. If messages are frequently exchanged across the network, the communication cost will easily dominate the computation cost.

There has also been a flurry of research studying the implementation of stochastic algorithms in the fully distributed setting~\cite{zinkevich2010parallelized,zhang2012communication,newman2007distributed, gemulla2011large,liu2010distributed}. Although promising results have been reported, the implementations proposed to date have their limitations---they have been designed for specific algorithms, or they require careful partitioning of the data to avoid inconsistency.

In this paper, we propose a general framework for parallelizing stochastic algorithms on multi-node distributed systems. Our framework is called \splash\ ({\bf S}ystem for {\bf P}arallelizing {\bf L}earning {\bf A}lgorithms with {\bf S}tochastic Met{\bf h}ods). Splash consists of a programming interface and an execution engine. Using the programming interface, the user develops sequential stochastic algorithms without thinking about issues of distributed computing. The algorithm is then automatically parallelized by the execution engine. The parallelization is communication efficient, meaning that its separate threads don't communicate with each other until all of them have processed a large bulk of data. Thus, the inter-node communication need not be a performance bottleneck.

\paragraph{Programming Interface} The programming interface is designed around a key paradigm: implementing incremental updates that processes weighted data. Unlike existing distributed machine learning systems~\cite{dean2012large,xing2013petuum,li2014scaling,murray2013naiad} which requires the user to explicitly specify a distributed algorithm, Splash asks the user to implement a processing function that takes an individual data element as input to incrementally update the corresponding variables. When this function is iteratively called on a sequence of samples, it defines a sequential stochastic algorithm. It can also be called in a distributed manner for constructing parallel algorithms, which is the job of the execution engine. This programming paradigm  allows one algorithmic module working on different computing environments, no matter if it is a single-core processor or a large-scale cluster. As a consequence, the challenge of parallelizing these algorithms has been transferred from the developer side to the system side.

To ensure parallelizability, the user is asked to implement a slightly stronger version of the base sequential algorithm: it needs to be capable of processing \emph{weighted samples}. An $m$-weighted sample tells the processing function that the sample appears $m$ times consecutively in the sequence.
Many stochastic algorithms can be generalized to processing weighted samples without sacrificing computational efficiency. We will demonstrate SGD and collapsed Gibbs sampling as two concrete examples. Since the processing of weighted samples can be carried out within a sequential paradigm, this requirement does not force the user to think about a distributed implementation. 

\paragraph{Execution Engine} In order to parallelize the algorithm, Splash converts a distributed processing task into a sequential processing task using distributed versions of \emph{averaging} and \emph{reweighting}. During the execution of the algorithm, we let every thread sequentially process its local data. The local updates are iteratively averaged to construct the global update. Critically, however, although averaging reduces the variance of the local updates, it doesn't reduce their bias. In contrast to the sequential case in which a thread processes a full sequence of random samples, in the distributed setting every individual thread touches only a small subset of samples, resulting in a significant bias relative to the full update. Our reweighting scheme addresses this problem by feeding the algorithm with weighted samples, ensuring that the total weight processed by each thread is equal to the number of samples in the full sequence. This helps individual threads to generate nearly-unbiased estimates of the full update. Using this approach, \splash\ automatically detects the best degree of parallelism for the algorithm. \\

Theoretically, we prove that \splash\ achieves the optimal rate of convergence for parallelizing SGD, assuming that the objective function is smooth and strongly convex. We conduct extensive experiments on a variety of stochastic algorithms, including algorithms for logistic regression, collaborative filtering and topic modeling. The experiments verify that \splash\ can yield orders-of-magnitude speedups over single-thread stochastic algorithms and over state-of-the-art batch algorithms.

Besides its performance, \splash\ is a contribution on the distributed computing systems front, providing a flexible interface for the implementation of stochastic algorithms. We build \splash\ on top of Apache Spark~\cite{zaharia2012resilient}, a popular distributed data-processing framework for batch algorithms. \splash\ takes the standard Resilient Distributed Dataset (RDD) of Spark as input and generates an RDD as output. The data structure also supports default RDD operators such as Map and Reduce, ensuring convenient interaction with Spark. Because of this integration, \splash\ works seamlessly with other data analytics tools in the Spark ecosystem, enabling a single system to address the entire analytics pipeline. 

\vspace{-5pt}
\section{Shared and Local Variables}
\label{sec:background}

In this paper, we focus on the stochastic algorithms which take the following general form.
At step~$t$, the algorithm receives a data element $z_t$ and a vector of \emph{shared variables} 
$v_t$. Based on these values the algorithm performs an incremental update $\Delta(z_t,v_t)$ 
on the shared variable:
\begin{align}\label{eqn:incremental-update}
	v_{t+1}\leftarrow v_t + \Delta(z_t,v_t)
\end{align}
For example, stochastic gradient descent (SGD) fits this general framework.  Letting $x$ denote
a random data element $x$ and letting $w$ denote a parameter vector, SGD performs the update:
\begin{align}
	t \leftarrow t+1 \quad \mbox{and} \quad w \leftarrow w - \eta_t \nabla \ell(w;x)\label{eqn:sgd-update-w}
\end{align}
where $\ell(\cdot;x)$ is the loss function associated with the element and $\eta_t$ is the stepsize at time $t$. In this case both $w$ and $t$ are shared variables. 

There are several stochastic algorithms using \emph{local variables} in their computation. Every local variable is associated with a specific data element. For example, the collapsed Gibbs sampling algorithm for LDA~\cite{griffiths2004finding} maintains a topic assignment for each word in the corpus. Suppose that a topic $k\in \{1,\dots,K\}$ has been sampled for a word $w$, which is in document $d$. The collapsed Gibbs sampling algorithm updates the word-topic counter $n_{wk}$ and the document-topic counter $n_{dk}$ by
\begin{align}\label{eqn:gibbs-first-update}
	n_{wk} \leftarrow n_{wk} + 1 \quad \mbox{and} \quad n_{dk} \leftarrow n_{dk} + 1.
\end{align}
The algorithm iteratively resample topics for every word until the model parameters converge. When a new topic is sampled for the word $w$, the following operation removes the old topic before drawing the new one:
\begin{align}\label{eqn:gibbs-second-update}
	n_{wk} \leftarrow n_{wk} - 1 \quad \mbox{and} \quad n_{dk} \leftarrow n_{dk} - 1.
\end{align}
Update~\eqref{eqn:gibbs-first-update} and update~\eqref{eqn:gibbs-second-update} are executed at different stages of the algorithm but they share the same topic $k$. Thus, there should be a local variable associated with the word $w$ storing the topic. \splash\ supports creating and updating local variables during the algorithm execution.

The usage of local variables can sometimes be tricky. Since the system carries out automatic reweighting and rescaling (refer to Section~\ref{sec:strategy}), any improper usage of the local variable may cause inconsistent scaling issues. The system thus provides a more robust interface called ``delayed operator'' which substitutes the functionality of local variables in many situations. In particular, the user can declare an operation such as~\eqref{eqn:gibbs-second-update} as a delayed operation and suspend its execution to the next time when the same element is processed. The scaling consistency of the delay operation is guaranteed by the system.

Shared variables and local variables are stored separately. In particular, shared variables are replicated on every data partition. Their values are synchronized. The local variables, in contrast, are stored with the associated data elements and will never be synchronized. This storage scheme optimizes the communication efficiency and allows for convenient element-wise operations.

\section{Programming with Splash}
\label{sec:programming-interface}

\splash\ allows the user to write self-contained Scala applications using its programming interface. The goal of the programming interface is to make distributed computing transparent to the user. \splash\ extends Apache Spark to provide an abstraction called a \emph{Parametrized RDD} for storing and maintaining the distributed dataset. The Parametrized RDD is based on the Resilient Distributed Dataset (RDD)~\cite{zaharia2012resilient} used by Apache Spark. It can be created from a standard RDD object:
\[
	\tt val~paramRdd = new~ParametrizedRDD(rdd).
\]
We provide a rich collection of interfaces to convert the components of Parametrized RDD to standard RDDs, facilitating the interaction between Splash and Spark.
%The $\tt paramRdd$ object contains the same number of partitions as the original $\tt rdd$ object. Each element of $\tt paramRdd$ comes from an element of $\tt rdd$, but with two additional features: (1) each element of $\tt paramRdd$ is equipped with a $\tt localVar$ object, which maintains the local variables associated with the element; (2) each partition of $\tt paramRdd$ maintains with a $\tt sharedVar$ object, which is the set of shared variables. 
To run algorithms on the Parametrized RDD, the user creates a function called $\tt process$ which implements the stochastic algorithm, then calls the method
\[
	\tt paramRdd.run(process)
\]
to start running the algorithm. In the default setting, the execution engine takes a full pass over the dataset by calling $\tt run()$ once. This is called one \emph{iteration} of the algorithm execution. The inter-node communication occurs only at the end of the iteration.
The user may call $\tt run()$ multiple times to take multiple passes over the dataset. 

The $\tt process$ function is implemented using the following format:
\[
	\tt def~process(elem: Any,~weight: Int,~sharedVar : VarSet,~localVar : VarSet) \{\dots\}
\]
It takes four arguments as input: a single element $\tt elem$, the weight of the element, the shared variable $\tt sharedVar$ and the local variable $\tt localVar$ associated with the element. The goal is to update $\tt sharedVar$ and $\tt localVar$ according to the input. 

\splash\ provides multiple ways to manipulate these variables. Both local and shared variables are manipulated as key-value pairs. The key must be a string; the value can be either a real number or an array of real numbers. Inside the $\tt process$ implementation, the value of local or shared variables can be accessed by $\tt localVar.get(key)$ or $\tt sharedVar.get(key)$. The local variable can be updated by setting a new value for it: 
\[
	\tt localVar.set(key, value)
\]
The shared variable is updated by \emph{operators}. For example, using the \emph{add} operator, the expression
\begin{align*}
	\tt sharedVar.add(key, delta)
\end{align*}
adds a scalar $\tt delta$ to the variable. The SGD updates~\eqref{eqn:sgd-update-w} can be implemented via several add operators. Other operators supported by the programming interface, including \emph{delayed add} and \emph{multiply}, are introduced in Section~\ref{sec:strategy}.
Similar to the standard RDD, the user can perform $\tt map$ and $\tt reduce$ operations directly on the Parametrized RDD. For example, after the algorithm terminates, the expression
\begin{align*}
	\tt val~loss = paramRdd.map(evalLoss).sum()
\end{align*}
evaluates the element-wise losses and aggregates them across the dataset.

%Since \splash\ utilizes the job scheduler of Apache Spark, it allows us to implement the framework in less than 1,300 lines of code. We briefly describe the workflow of the execution engine of \splash. Once the user calls $\tt paramRdd.run()$, the master node picks a random set of active workers. The number of active workers is by default equal to the number of partitions in $\tt paramRdd$. For each partition, the data elements are first randomly permuted, then sequentially processed by the user-specified $\tt process$ function. The element weights are automatically determined by the system following the algorithm described in Appendix~\ref{sec:determine-weight}.
%
%Once every worker finishes their jobs, the execution engine calls a $\tt reduce$ operator of Spark to aggregate local updates. The global update is then broadcasted to all workers, making the shared variables synchronized across all partitions. Although \splash\ is a synchronous system, it performs infrequent synchronization, so that the communication cost is not a major contributor to the overall running time. It is worth noting that the system only synchronizes shared variables that have been modified in the current iteration. Thus, those infrequently updated shared variables won't be a bottleneck of the communication efficiency.

\section{Strategy for Parallelization}
\label{sec:splash-strategy}

In this section, we first discuss two naive strategies for parallelizing a stochastic algorithm and their respective limitations. These limitations motivate the strategy that \splash\ employs.

\subsection{Two naive strategies}
\label{sec:simple-strategy}
We denote by $\Delta(\mathcal{S})$ the incremental update on variable $v$ after processing the set of samples~$\mathcal{S}$. Suppose that there are $m$ threads and each thread processes a subset $S_i$ of $\mathcal{S}$. If the $i$-th thread increments the shared variable by $\Delta(S_i)$, then the \emph{accumulation scheme} constructs a global update by accumulating local updates:
\begin{align}\label{eqn:accumulate-update}
	v_{\rm new} = v_{\rm old} + \sum_{i=1}^m \Delta(S_i).
\end{align}
The scheme~\eqref{eqn:accumulate-update} provides a good approximation to the full  update if the batch size $|D_i|$ is sufficiently small~\cite{agarwal2011distributed}. However, frequent communication is necessary to ensure a small batch size. For distributed systems connected by commodity networks, frequent communication is prohibitively expensive, even if the communication is asynchronous.

Applying scheme~\eqref{eqn:accumulate-update} on a large batch may easily lead to divergence. Taking SGD as an example: if all threads starts from the same vector $w_{\rm old}$, then after processing a large batch, the new vector on each thread will be close to the optimal solution $w^*$. If the variable is updated by formula~\eqref{eqn:accumulate-update}, then we have
\begin{align*}
	w_{\rm new} - w^*  &=  w_{\rm old} - w^* + \sum_{i=1}^m \Delta(S_i) \approx  w_{\rm old} - w^* + \sum_{i=1}^m (w^* -w_{\rm old}) = (m-1)(w^* - w_{\rm old}).
\end{align*}
Clearly SGD will diverge if $m\geq 3$. 

One way to avoid divergence is to multiply the incremental change by a small coefficient.
When the coefficient is $1/m$, the variable is updated by
\begin{align}\label{eqn:damped-update}
	v_{\rm new} = v_{\rm old} + \frac{1}{m} \sum_{i=1}^m  \Delta(S_i).
\end{align}
This \emph{averaging scheme} usually avoids divergence. However, since the local updates are computed on $1/m^{th}$ of $\mathcal{S}$, they make little progress comparing to the full sequential update. Thus the algorithm converges substantially slower than its sequential counterpart after processing the same amount of data. See Section~\ref{sec:toy-example} for an empirical evidence of this claim.

\subsection{Our strategy}
\label{sec:strategy}

We now turn to describe the \splash\ strategy for combining parallel updates. First we introduce the operators that \splash\ supports for manipulating shared variables. Then we illustrate how conflicting updates are combined by the reweighting scheme. 

\paragraph{Operators} The programming interface  allows the user to manipulate shared variables inside their algorithm implementation via \emph{operators}. An operator is a function that maps a real number to another real number. \splash\ supports three types of operators: \emph{add}, \emph{delayed add} and \emph{multiply}. The system employs different strategies for parallelizing different types of operators. 

The \emph{add} operator is the the most commonly used operator. When the operation is performed on variable $v$, the variable is updated by
$v \leftarrow v + \delta$ where $\delta$ is a user-specified scalar. The SGD update~\eqref{eqn:sgd-update-w} can be implemented using this operator. 

The \emph{delayed add} operator performs the same mapping $v \leftarrow v + \delta$; however, the operation will not be executed until the next time that the same element is processed by the system. Delayed operations are useful in implementing sampling-based stochastic algorithms. In particular, before the new value is sampled, the old value should be removed. This ``reverse'' operation can be declared as a delayed operator when the old value was sampled, and executed before the new value is sampled. See Section~\ref{sec:generalize-stochastic-algorithm} for a concrete example.

The \emph{multiply} operator scales the variable by $v \leftarrow \gamma \cdot v$ where $\gamma$ is a user-specified scalar. The multiply operator is especially efficient for scaling high-dimensional arrays. The array multiplication costs $\order(1)$ computation time, independent of the dimension of the array. 
The fast performance is achieved by a ``lazy update'' scheme. For every array $u$, there is a variable $V$ maintaining the product of all multipliers applied to the array.
The multiply operator updates $V\leftarrow \gamma\cdot V$ with $\order(1)$ time. For the $i$-th element $u_i$, a variable $V_i$ maintains the product of all multipliers applied to the element. When the element is accessed, the system updates $u_i$ and $V_i$ by
\[
	u_i \leftarrow \frac{V}{V_i}\cdot u_i \quad \mbox{and}
	\quad V_i \leftarrow V. 
\]
In other words, we delay the multiplication on individual element until it is used by the program. As a consequence, those infrequently used elements won't be a bottleneck on the algorithm's performance.
See Section~\ref{sec:generalize-stochastic-algorithm} for a concrete example.

\paragraph{Reweighting} Assume that there are $m$ thread running in parallel. Note that all \splash\ operators are linear transformations. When these operators are applied sequentially, they merge into a single linear transformation. Let $S_i$ be the sequence of samples processed by thread $i$, which is a fraction $1/m$ of the full sequence $\mathcal{S}$. For an arbitrary shared variable $v$, we can write thread $i$'s transformation of this variable in the following form:
\begin{align}\label{eqn:group-level-operator}
	v \leftarrow \Gamma(S_i) \cdot v + \Delta(S_i)  + T(S_i),
\end{align}
Here, both $\Gamma(S_i)$, $\Delta(S_i)$ and $T(S_i)$ are thread-level operators constructed by the execution engine: $\Gamma(S_i)$ is the aggregated multiply operator, $\Delta(S_i)$ is the term resulting from the add operators, and $T(S_i)$ is the term resulting from the delayed add operators executed in the current iteration. A detailed construction of $\Gamma(S_i)$, $\Delta(S_i)$ and $T(S_i)$ is given in  Appendix~\ref{sec:construct-group-level-transformation}.

As discussed in Section~\ref{sec:simple-strategy}, directly combining these transformations leads to divergence or slow convergence (or both). The reweighting scheme addresses this dilemma by assigning weights to the samples. Since the update~\eqref{eqn:group-level-operator} is constructed on a fraction $1/m$ of the full sequence $\mathcal{S}$, we reweight every element by $m$ in the local sequence. After reweighting, the data distribution of $S_i$ will approximate the data distribution of $\mathcal{S}$. If the update~\eqref{eqn:group-level-operator} is a (randomized) function of the data distribution of $S_i$, then it will approximate the full sequential update after the reweighting, thus generating a nearly unbiased update.

More concretely, the algorithm manipulates the variable by taking sample weights into account. An $m$-weighted sample tells the algorithm that it appears $m$ times consecutively in the sequence. We rename the transformations in~\eqref{eqn:group-level-operator} by $\Gamma(mS_i)$, $\Delta(mS_i)$ and $T(mS_i)$, emphasizing that they are constructed by processing $m$-weighted samples. Then we redefine the transformation of thread~$i$ by
\begin{align}\label{eqn:weighted-group-level-operator}
	v \leftarrow \Gamma(mS_i)\cdot v + \Delta(mS_i) + T(m S_i)
\end{align}
and define the global update by
\begin{align}\label{eqn:average-weighted-groups}
	v_{\rm new} = \frac{1}{m} \sum_{i=1}^m \Big( \Gamma(mG_i)\cdot v_{\rm old} + \Delta(mS_i) \Big) + \sum_{i=1}^m T(m S_i).
\end{align}
Equation~\eqref{eqn:average-weighted-groups} combines the transformations of all threads. The terms $\Gamma(mS_i)$ and $\Delta(mS_i)$ are scaled by a factor $1/m$ because they were constructed on $m$ times the amount of data. The term $T(mS_i)$ is not scaled, because the delayed operators were declared in earlier iterations, independent of the reweighting. Finally, the scaling factor $1/m$ should be multiplied to all delayed operators declared in the current iteration, because these delayed operators were also  constructed on $m$ times the amount of data. 

\paragraph{Determining the degree of parallelism} To determine the thread number $m$, the execution engine partitions the 
available cores into different-sized groups. Suppose that group $i$ contains $m_i$ cores. These cores will  execute the algorithm tentatively on $m_i$ parallel threads. The best thread number is then determined by cross-validation and is dynamically updated. The cross-validation requires the user to implement a loss function, which takes the variable set and an individual data element as input to return the loss value.  See Appendix~\ref{sec:determine-weight} for a detailed description. To find the best degree of parallelism, the base algorithm needs to be robust in terms of processing a wide range of sample weights.

\subsection{Generalizing stochastic algorithms} 
\label{sec:generalize-stochastic-algorithm}
Many stochastic algorithms can be generalized to processing weighted samples without sacrificing computational efficiency. The most straightforward generalization is to repeat the single-element update $m$ times. For example, one can generalize the SGD updates~\eqref{eqn:sgd-update-w} by
\begin{align}\label{eqn:weighted-sgd-update-w}
	t\leftarrow t + m \quad \mbox{and} \quad w \leftarrow w - \eta_{t,m} \nabla\ell(w;x)
\end{align}
where $\eta_{t,m}\defeq \sum_{i=t-m+1}^t \eta_i$ is the sum of all stepsizes in the time interval $[t-m+1,t]$, and $\eta_i$ is the stepsize for the unit-weight sequential SGD. If $m$ is large, computing $\eta_{t,m}$ might be expensive. We may approximate it by
\[
	\eta_{t,m} \approx \int_{t-m+1}^{t+1} \eta_z dz
\]
if the right-hand side has a closed-form solution, or simply approximate it by $\eta_{t,m} \approx m\eta_t$. 

In many applications, the loss function $\ell(w;x)$ can be decomposed as $\ell(w;x)\defeq f(w;x) + \frac{\lambda}{2}\ltwos{w}^2$ where the second term is the $\ell_2$-norm regularization. Thus, we
have $\nabla\ell(w;x) = \nabla f(w;x) + \lambda w$. If the feature vector $x$ is sparse, then $\nabla f(w;x)$ is usually sparse as well.
In this case, we have a more efficient implementation of~\eqref{eqn:weighted-sgd-update-w}:
\begin{align*}
	&t\leftarrow t + m \qquad &\mbox{(via add operator)}, \\
	&w \leftarrow (1 - \eta_{t,m} \lambda)\cdot w  
	&\mbox{(via multiply operator)},\\
	& w \leftarrow w - \eta_{t,m} \nabla f(w;x)  &\mbox{(via add operator)}.
\end{align*}
Note that the multiply operator has complexity~$\order(1)$. Thus, the overall complexity is proportional to the number of non-zero components of $\nabla f(w;x)$. 
If $\nabla f(w;x)$ is a sparse vector, then this update will be more efficient than~\eqref{eqn:weighted-sgd-update-w}. It demonstrates the benefit of combining different types of operators.

Note that equation~\eqref{eqn:weighted-sgd-update-w} scales the stepsize with respect to $m$, which might be unsafe $m$ is very large. Karampatziakis and Langford~\cite{karampatziakis2010online} propose a robust approach to dealing with large importance weights in SGD. The programming interface allows the user to implement the approach by 
Karampatziakis and Langford~\cite{karampatziakis2010online}.

We take the collpased Gibbs sampling algorithm for LDA as a second example. The algorithm iteratively draw a word $w$ from document $d$, and sample the topic of $w$ by
\begin{align}\label{eqn:lda-gibbs-sampling}
	P({\rm topic} = k | d,w) \propto \frac{(n_{dk} + \alpha)(n_{wk} + \beta)}{n_{k} + \beta W}.
\end{align}
Here, $W$ is the size of the vocabulary; $n_{dk}$ is the number of words in document $d$ that has been assigned topic $k$;
$n_{wk}$ is the total number of times that word $w$ is assigned to topic $k$ and $n_k := \sum_{w} n_{wk}$.
The constants $\alpha$ and $\beta$ are hyper-parameters of the LDA model. When a topic $k$ is sampled for the word, the algorithm updates  $n_{wk}$ and $n_{dk}$ by~\eqref{eqn:gibbs-first-update}. When a new topic will be sampled for the same word, the algorithm removes the old topic $k$ by~\eqref{eqn:gibbs-second-update}. If the current word has weight $m$, then we can implement the algorithm by
\begin{align}
	& n_{wk} \leftarrow n_{wk} + m \quad \mbox{and} \quad n_{dk} \leftarrow n_{dk} + m  &\mbox{(via add operator)},\label{eqn:general-gibbs-first}\\
	& n_{wk} \leftarrow n_{wk} - m \quad \mbox{and} \quad n_{dk} \leftarrow n_{dk} - m  &\mbox{(via delayed add operator)}.\label{eqn:general-gibbs-second}
\end{align}
As a consequence, the update~\eqref{eqn:general-gibbs-first} will be executed instantly. The update~\eqref{eqn:general-gibbs-second} will be executed at the next time when the same word is processed.

\subsection{A toy example}
\label{sec:toy-example}

\begin{figure}[t]
\centering
\includegraphics[width = 0.7\textwidth]{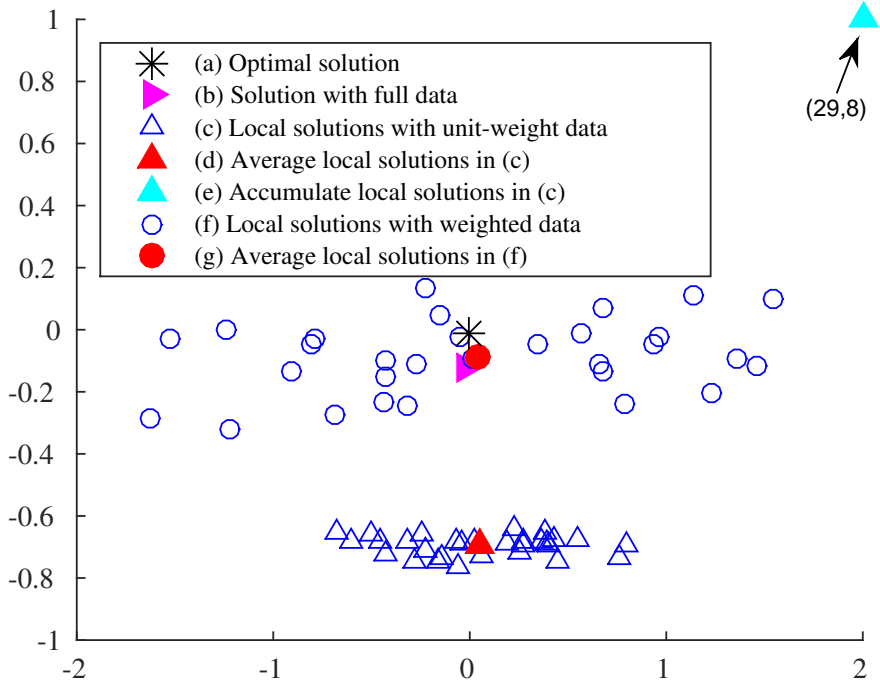}
\caption{Comparing parallelization schemes on a simple convex optimization problem. Totally $N=3,000$ samples are partitioned into $m = 30$ batches. Each batch is processed by an independent thread running stochastic gradient descent. Each thread uses either unit-weight data or weighted data (weight = 30). The local solutions are combined by either averaging or accumulation. From the plot, we find that combining weighted solutions achieves the best performance.}
\label{fig:example}
\end{figure}

We present a toy example illustrating the strategy described in Section~\ref{sec:strategy}. Consider the following convex optimization problem. There are $N=3,000$ two-dimensional vectors represented by $x_1,\dots,x_N$, such that $x_i$ is randomly and independently drawn from the normal distribution $x\sim N(0,I_{2\times 2})$. The goal is to find a two-dimensional vector $w$ which minimizes the weighted distance to all samples. More precisely, the loss function on sample $x_i$ is defined by
\[
	\ell(w;x_i) \defeq (x_i - w)^T \left( \begin{array}{cc}
	1 & 0\\
	0 & \frac{1}{100}
	\end{array}\right)(x_i-w)
\]
and the overall objective function is $L(w)\defeq \frac{1}{N}\sum_{i=1}^N \ell(w;x_i)$. We want to find the vector that minimizes the objective function $L(w)$.

We use the SGD update~\eqref{eqn:weighted-sgd-update-w} to solve the problem.
The algorithm is initialized by $w_0 = (-1,-1)^T$ and the stepsize is chosen by $\eta_t = 1/\sqrt{t}$.
For parallel execution, the dataset is evenly partitioned into $m=30$ disjoint subsets, such that each thread accesses to a single subset, containing $1/30$ faction of data.
The sequential implementation and the parallel implementations are compared in Figure~\ref{fig:example}. Specifically, we compare seven types of implementations defined by different strategies:

\begin{enumerate}[(a)]
\item The exact minimizer of $L(w)$.
\item The solution of SGD achieved by taking a full pass over the dataset. The dataset contains $N = 3,000$ samples.
\item The local solutions by $30$ parallel threads. Each thread runs SGD by taking one pass over its local data. The local dataset contains $100$ samples.
\item Averaging local solutions in (c). This is the averaging scheme described by formula~\eqref{eqn:damped-update}.
\item Aggregating local solutions in (c). This is the accumulation scheme described by formula~\eqref{eqn:accumulate-update}.
\item The local solution by $30$ parallel threads processing weighted data. Each element is weighted by $30$. Each thread runs  SGD by taking one pass over its local data. 
\item Combining parallel updates by formula~\eqref{eqn:average-weighted-groups}, setting sample weight $m = 30$. Under this setting, formula~\eqref{eqn:average-weighted-groups} is equivalent to averaging local solutions in~(f).
\end{enumerate}

In Figure~\ref{fig:example}, we observe that solution (b) and solution (g) achieve the best performance. Solution (b) is obtained by a sequential implementation of SGD: it is the baseline solution that parallel algorithms target at approaching.
Solution (g) is obtained by \splash\ with the reweighting scheme. The solutions obtained by other parallelization schemes, namely solution (d) and (e), have poor performances. In particular, the averaging scheme (d) has a large bias relative to the optimal solution. The accumulation scheme (e) diverges far apart from the optimal solution. 

To see why \splash\ is better, we compare local solutions (c) and (f). They correspond to the unweighted SGD and the weighted SGD  respectively. We find that solutions (c) have a significant bias but relatively small variance. In contrast, solutions (f) have greater variance but much smaller bias. It verifies our intuition that reweighting helps to reduce the bias by enlarging the local dataset. Note that averaging reduces the variance but doesn't change the bias. It explains why averaging works better with reweighting.

\section{Convergence Analysis}
\label{sec:convergence}

In this section, we study the SGD convergence when it is parallelized by \splash. 
The goal of SGD is to minimize an empirical risk function
\[
L(w) = \frac{1}{|S|}\sum_{x\in S} \ell(w;x),
\]
where $S$ is a fixed dataset and $w\in \R^d$ is the vector to be minimized over. Suppose that there are $m$ threads running in parallel. At every iteration, thread $i$ randomly draws (with replacement) a subset of samples $S_i$  of length $n$ from the dataset~$S$. The thread sequentially processes $S_i$ by SGD. The per-iteration update is
\begin{align}\label{eqn:sgd-with-projection}
	t&\leftarrow t + m \quad \mbox{and} \quad w\leftarrow w + \big( \Pi_W(w - m \eta_t \nabla\ell(w;x) ) - w \big),
\end{align}
where the sample weight is equal to $m$. We have generalized the update~\eqref{eqn:weighted-sgd-update-w} by introducing $\Pi_W(\cdot)$ as a projector to a feasible set $W$ of the vector space. Projecting to the feasible set is a standard post-processing step for an SGD iterate. At the end of the iteration, updates are synchronized by equation~\eqref{eqn:average-weighted-groups}. This is equivalent to computing:
\begin{align}\label{eqn:fix-group-number-sgd-combine}
	t_{\rm new} = t_{\rm old} + mn \quad \mbox{and}\quad w_{\rm new} = \frac{1}{m} \sum_{i=1}^m \Big( w_{\rm old} + \Delta(mD_i)\Big).
\end{align}
We denote by $\wstar \defeq \arg\min_{w\in W}L(w)$ the minimizer of the objective function, and denote by $w^T$ the combined vector after the $T$-th iteration.

\paragraph{General convex function}

For general convex functions, we start by introducing three additional terms. Let $w_{i,j}^k$ be the value of vector $w$ at iteration~$k$, when thread $i$ is processing the $j$-th element of $S_i$. Let $\eta_{i,j}^k$ be the stepsize associated with that update. We define a weighted average vector:
\[
	\wbar^T = \frac{\sum_{k=1}^T \sum_{i=1}^m \sum_{j=1}^n \eta_{i,j}^k w_{i,j}^k}{\sum_{k=1}^T \sum_{i=1}^m \sum_{j=1}^n \eta_{i,j}^k}.
\]
Note that $\wbar^T$ can be computed together with $w^T$. For general convex $L$, the function value $L(\wbar^T)$ converges to $L(\wstar)$. See Appendix~\ref{sec:proof-global-convergence} for the proof.

\begin{theorem}\label{theorem:global-convergence}
Assume that $\ltwos{\nabla \ell(w;x)}$ is bounded for all $(w,x)\in W\times S$. Also assume that $\eta_t$ is a monotonically decreasing function of $t$ such that $\sum_{t=1}^\infty \eta_t = \infty$ and $\sum_{t=1}^\infty \eta_t^2 < \infty$. Then we have
\[
	\lim_{T\to \infty } \E[L(\wbar^T) - L(w^*)] = 0.
\]
\end{theorem}

%For sequential execution (when $m=1$), Theorem~\ref{theorem:global-convergence} is a classicial result for SGD. 
%We have shown that \splash\ converges in the parallel setting as well. Although the theorem itself doesn't imply that \splash\ achieves any speedup, the empirical result shows that it does converge much faster than the single-thread SGD.

\paragraph{Smooth and strongly convex function}

We now turn to study smooth and strongly convex objective functions. We make three assumptions on the objective function.
Assumption~\ref{assumption:parameter-space} restricts the optimization problem in a bounded convex set. Assumption~\ref{assumption:smoothness-sgd} and Assumption~\ref{assumption:strong-convexity-sgd} 
require the objective function to be sufficiently smooth and strongly convex in that set. 

\vspace{5pt}
\begin{assumption}
\label{assumption:parameter-space}
The feasible set $W \subset \R^d$ is a compact convex
set of finite diameter $R$. Moreover, $\wstar$ is an interior point of $W$; i.e.,
there is a set $U_\rho \defeq \{w\in\R^d:\ltwo{w-\wstar} <  \rho\}$  such that $U_\rho \subset W$.
\end{assumption}

\begin{assumption}
  \label{assumption:smoothness-sgd}
  There are finite constants $\liphessian$, $\lipobj$ and $\lipgrad$ such that
    $\ltwos{\nabla^2 L(w; x) - \nabla^2
      \ell(\wstar; x)} \le \liphessian
    \ltwos{w - \wstar}$, $\ltwos{\nabla \ell(w; x)} \le \lipobj$ and $ \ltwos{\nabla^2 \ell(w; x)}
    \le \lipgrad$
  for all $(w,x)\in W\times S$.
\end{assumption}

\begin{assumption}
\label{assumption:strong-convexity-sgd}
  The objective function $L$ is $\strongparam$-strongly
  convex over the space $W$, meaning that $\nabla^2 L(w) \succeq \strongparam I_{d \times d}$ for all $w \in W$.
\end{assumption}

As a preprocessing step, we construct an Euclidean ball $B$ of diameter $D \defeq \frac{\lambda}{4(L+G/\rho^2)}$ which contains the optimal solution $\wstar$. The ball center can be found by running the sequential SGD for a constant number of steps. During the \splash\ execution, if the combined vector $w^T \notin B$, then we project it to~$B$, ensuring that the distance between $w^T$ and $\wstar$ is bounded by $D$.  Introducing this projection step simplifies the theoretical analysis, but it may not be necessary in practice.

Under these assumptions, we provide an upper bound on the mean-squared error of $w^T$. The following theorem shows that the mean-square error decays as $1/(Tmn)$, inversely proportionally to the total number of processed samples. It is the optimal rate of convergence among all optimization algorithms which relies on noisy gradients~\cite{rakhlin2011making}. See Appendix~\ref{sec:proof-strong-convex-sgd} for the proof.

\vspace{5pt}
\begin{theorem}
\label{theorem:strong-convex-sgd}
Under Assumptions~\ref{assumption:parameter-space}-\ref{assumption:strong-convexity-sgd}, if we choose the stepsize $\eta_t = \frac{2}{\lambda t}$, then
the output $w^T$ has mean-squared error:
  \begin{equation}
    \E\left[\ltwos{w^T - \wstar}^2\right]
    \leq
    \frac{4 G^2}{\strongparam^2 T m n} + \frac{C_1}{T m^{1/2}n^{3/2}}
    + \frac{C_2}{T n^2},
    \label{eqn:sgd-bound}
  \end{equation}
 where $C_1$ and $C_2$ are constants independent of $T$, $m$ and $n$.
\end{theorem}

When the local sample size $n$ is sufficiently larger than the thread number $m$ (which is typically true), the last two terms on the right-hand side of bound~\eqref{eqn:sgd-bound} are negligibly small. Thus, the mean-squared error is dominated by the first term, which scales as $1/(T m n)$. 

\section{Experiments}
\label{sec:experiment}

In this section, we report the empirical performance of \splash\ on three machine learning tasks: logistic regression, collaborative filtering and topic modeling. Our implementation of \splash\ runs on an Amazon EC2 cluster with eight nodes. Each node is powered by an eight-core Intel Xeon E5-2665 with $30$GB of memory and was connected to a commodity 1GB network, so that the cluster contains 64 cores. 
For all experiments, we compare \splash\ with MLlib v1.3~\cite{meng2015mllib} --- the official distributed machine learning library for Spark. We also compare \splash\ against single-thread stochastic algorithms.

\subsection{Logistic Regression}

We solve a digit recognition problem on the MNIST 8M dataset~\cite{loosli2007training} using multi-class logistic regression. The dataset contains 8 million hand-written digits. Each digit is represented by a feature vector of dimension $d = 784$. 
There are ten classes representing the digits 0-9. The goal is to minimize the following objective function:
\[
	L(w) \defeq \frac{1}{n}\sum_{i=1}^n - \langle w_{y_i}, x_i\rangle + \log\Big(\sum_{k=0}^{9} \exp^{\langle w_k, x_i\rangle}\Big)
\]
where $x_i\in \R^d$ is the feature vector of the $i$-th element and $y_i\in\{0,\dots,9\}$ is its label. The vectors $w_0,\dots,w_9\in \R^d$ are parameters of the logistic regression model.

\splash\ solves the optimization problem by SGD. We use equation~\eqref{eqn:weighted-sgd-update-w} to generalize SGD to processing weighted samples (the stepsize $\eta_{t,m}$ is approximated by $m\eta_t$). The stepsize $\eta_t$ is determined by the adaptive subgradient method (AdaGrad)~\cite{duchi2011adaptive}. We compare Splash against the single-thread SGD (with AdaGrad) and the MLlib implementation of L-BFGS~\cite{NocedalWrightbook}. Note that MLlib also provides a mini-batch SGD method, but in practice we found it converging substantially slower than L-BFGS.

\begin{figure}
\centering
\begin{tabular}{cc}
\includegraphics[height = 0.3\textwidth]{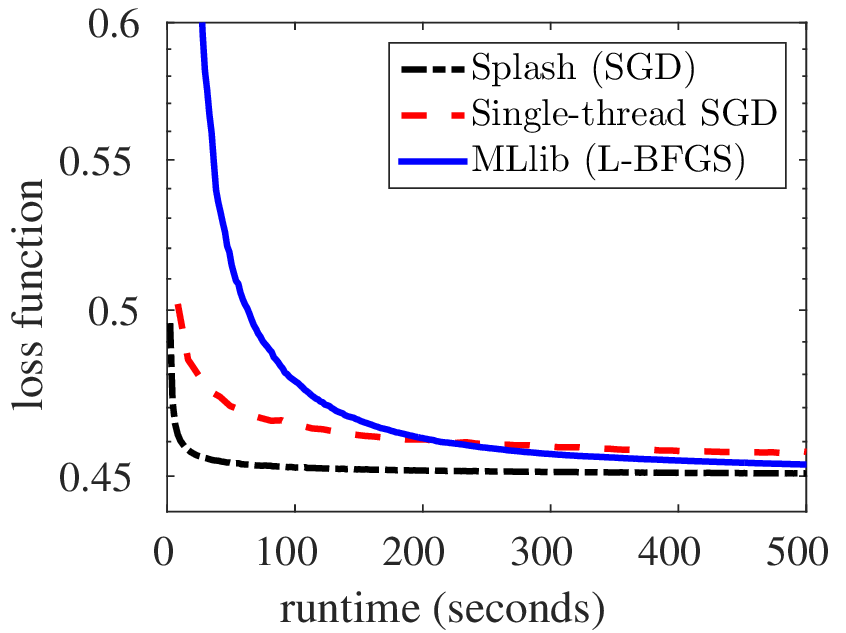}&
\includegraphics[height = 0.3\textwidth]{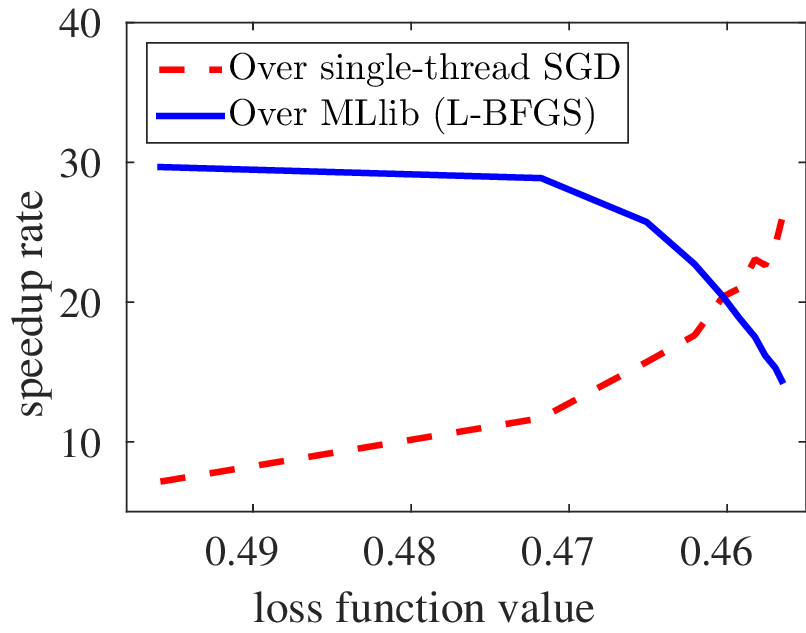}\\
(a) Loss function convergence & (b) \splash\ speedup rates
\end{tabular}
\caption{Multi-class logistic regression on the MNIST 8M digit recognition dataset. (a) The convergence of different methods; (b) The speedup over other methods for achieving the same loss function value.}\label{fig:mnist8m}
\end{figure}

Figure~\ref{fig:mnist8m}(a) shows the convergence plots of the three methods. \splash\ converges in a few seconds to a good solution. The single-thread AdaGrad and the L-BFGS algorithm converges to the same accuracy in much longer time. Figure~\ref{fig:mnist8m}(b)
demonstrates Splash's speedup over other methods. When the target loss decreases, the speedup rate over the single-thread SGD grows larger, while the speedup rate over MLlib drops lower. Thus, \splash\ is 15x - 30x faster than MLlib. Note that Splash runs a stochastic algorithm and L-BFGS is a batch method. It highlights the advantage of the stochastic method in processing large dataset.

\subsection{Collaborative Filtering}

We now turn to a personalized movie recommendation task. For this task, we use the Netflix prize dataset~\cite{bennett2007netflix}, which contains 100 million movie ratings made by 480k users on 17k movies. We split the dataset randomly into a training set and a test set, which contains 90\% and 10\% of the ratings respectively. 
The goal is to predict the ratings in the test set given ratings in the training set. 

The problem can be solved using collaborative filtering. Assume that each user $i$ is associated with a latent vector $u_i\in \R^d$, and each movie $j$ is associated with a latent vector $v_j\in \R^d$. The affinity score between the user and the movie is measure by the inner product $\langle u_i, v_j \rangle$. Given ratings in the training set, we define the objective function by:
\begin{align}\label{eqn:ef-loss}
	L(\{u_i\},\{v_j\}) \defeq \sum_{(i,j,r_{ij})\in S} \left( (\langle u_i, v_j \rangle - r_{ij})^2 + \lambda\ltwos{u_i}^2 + \lambda\ltwos{v_j}^2\right),
\end{align}
where $\mathcal{S}$ represents the training set; The triplet $(i,j,r_{ij})$ represents that the user $i$ gives rating $r_{ij}$ to the movie $j$.
In the training phase, we fit the user vectors $\{u_j\}$ and the movie vectors $\{v_j\}$ by minimizing~\eqref{eqn:ef-loss}. In the testing phase, we predict the ratings of a user $i$ which might not be in the training set. Let
$\{r_{ij}\}_{j\in J}$ be the observed ratings from the user, we compute the user vector $u_i$ by
\[
	u_i \defeq \arg\min_{u\in \R^d} \sum_{j\in J}(\langle u, v_j \rangle - r_{ij})^2 + \lambda\ltwos{u}^2,
\]
then predict the ratings on other movies by $\langle u_i, v_j \rangle$. The prediction loss is measured by the mean-squared error.

To minimize the objective function~\eqref{eqn:ef-loss}, we employ a SGD method. Let $I$ be the set of users. Let $J_i$ represents the set of movies that user $i$ has rated in the training set. We define an objective function with respect to $\{v_j\}$ as:
\begin{align}
	L(\{v_j\}) &\defeq \min_{\{u_i\}} L(\{u_i\},\{v_j\})\nonumber\\
	& = \sum_{i\in I} \Bigg( \min_{u\in \R^d} \Big\{ \sum_{j\in J_i} (\langle u, v_j \rangle - r_{ij})^2 + \lambda\ltwos{u}^2 \Big\} + \lambda \sum_{j\in J_i}  \ltwos{v_j}^2 \Bigg),\label{eqn:ef-marginal-loss}
\end{align}
so that the movie vectors are obtained by minimizing~\eqref{eqn:ef-marginal-loss}. SGD suffices to solve this problem because the objective function is a sum of individual losses.

In practice, we choose the dimension $d=100$ and regularization parameter $\lambda = 0.02$. Thus the number of parameters to be learned is 65 million. The movie vectors are initialized by a random unit vector on the ``first quadrant'' (all coordinates are positive). Splash runs the generalized SGD algorithm~\eqref{eqn:weighted-sgd-update-w} with stepsizes determined by AdaGrad. We compare Splash against the single-thread SGD method and the MLlib implementation of alternating least square (ALS) method. 
The ALS method minimizes the objective function~\eqref{eqn:ef-loss} by alternating minimization with respect to $\{u_i\}$ and with respect to $\{v_j\}$. 

\begin{figure}
\centering
\includegraphics[width = 0.4\textwidth]{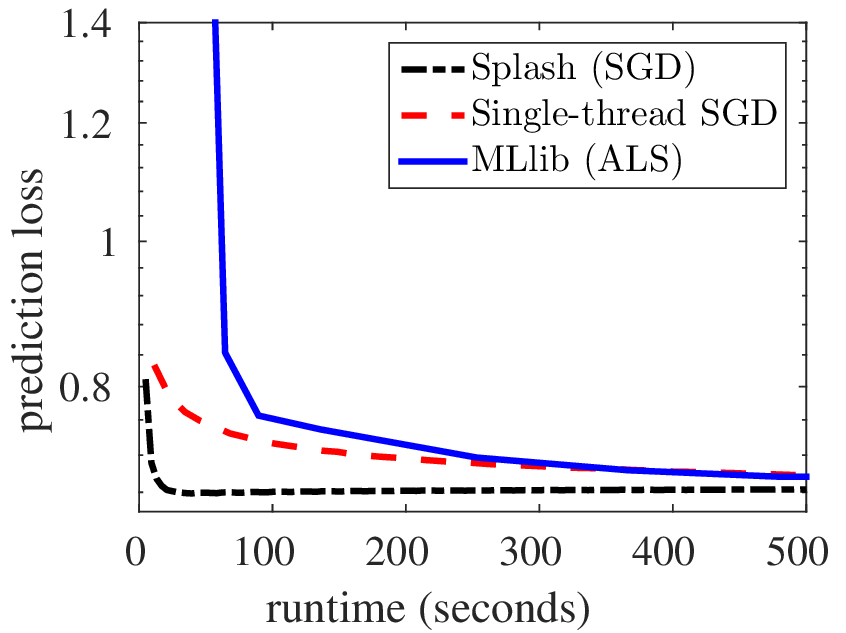}
\caption{Collaborative filtering on the Netflix prize dataset.}\label{fig:netflix}
\end{figure}

According to Figure~\ref{fig:netflix}, \splash\ converges much faster than the single-thread SGD and the ALS. This is because that SGD can learn accurate movie vectors by processing a fraction of the the data. For example, to achieve a prediction loss lower than $0.70$, it takes \splash\ only 13 seconds, processing 60\% of the training set. To achieve the same prediction loss, it takes the ALS 480 seconds, taking 40 passes over the full training set. In other words, \splash\ features a 36x speedup over the MLlib.

\subsection{Topic Modeling}

We use the NYTimes article dataset from the UCI machine learning repository~\cite{Lichman2013}. The dataset contains 300k documents and 100 million word tokens. The vocabulary size is 100k. The goal is to learn $K = 500$ topics from these documents. Each topic is represented by a multinomial distribution of words. The number of parameters to be learned is 200 million.

We employ the LDA model~\cite{blei2003latent} and choose hyper-parameters $\alpha = \beta = 0.1$. \splash\ runs the generalized collapsed Gibbs sampling algorithm~\eqref{eqn:general-gibbs-first}-\eqref{eqn:general-gibbs-second}. We also use the over-sampling technique~\cite{zhao2014same}, that is, for each word the algorithm independently samples 10 topics, each topic carrying 1/10 of the word's weight. We compare \splash\ with the single-thread collapsed Gibbs sampling algorithm and the MLlib implementation of the variational inference (VI) method~\cite{blei2003latent}. 

To evaluate the algorithm's performance, we resort to the predictive log-likelihood metric by Hoffman et al.~\cite{hoffman2013stochastic}. In particular, we partition the dataset into a training set $\mathcal{S}$ and a test set $\mathcal{T}$. The test set contains 10k documents. For each test document in $\mathcal{T}$, we partition its words into a set of observed words ${\bf w}_{\rm obs}$ and
held-out words ${\bf w}_{\rm ho}$, keeping the sets of unique words in ${\bf w}_{\rm obs}$ and ${\bf w}_{\rm ho}$ disjoint. We learn the topics from the training data $\mathcal{S}$, and then use that knowledge and the word set ${\bf w}_{\rm obs}$ to estimate the topic distribution for the test documents. Finally, the predictive log-likelihood of the held-out words, namely
$\log p({\bf w}_{\rm new} | {\bf w}_{\rm obs}, \mathcal{S})$, are computed. The performance of the algorithm is measured by the average predictive log-likelihood per held-out word.

\begin{figure}
\centering
\includegraphics[width = 0.4\textwidth]{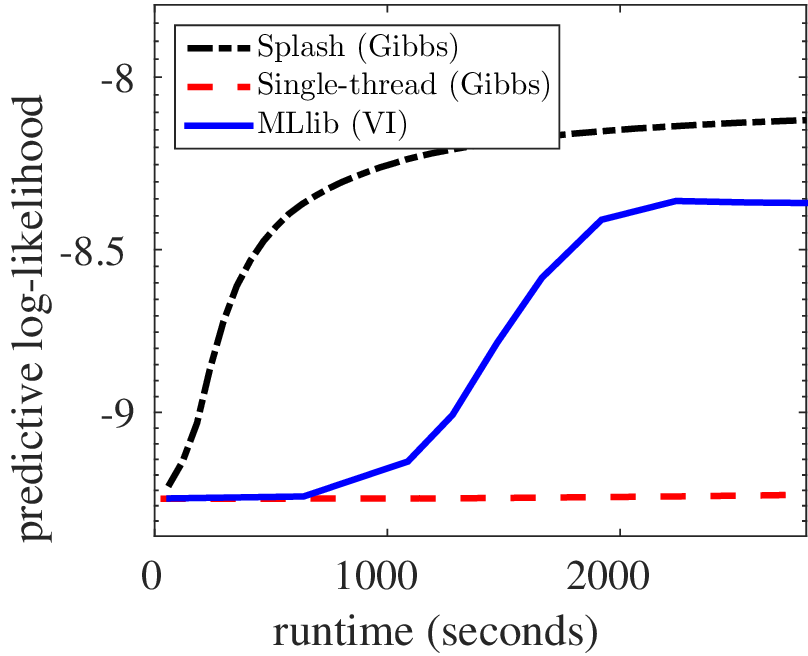}
\caption{Topic modeling on the NYTimes dataset. The LDA model learns $K = 500$ topics.}\label{fig:nytimes}
\end{figure}

Figure~\ref{fig:nytimes} plots the predictive log-likelihoods. Among the three methods, the single-thread collapsed Gibbs sampling algorithm exhibits little progress in the first 3,000 seconds. But when the algorithm is parallelized by Splash, it converges faster and better than the MLlib implementation of variational inference (VI). In particular, Splash converges to a predictive log-likelihoods of -8.12, while MLlib converges to -8.36. When measured at fixed target scores, Splash is 3x - 6x faster than MLlib.

\subsection{Runtime Analysis}

The runtime of a distributed algorithm can be decomposed into three parts: the computation time, the waiting time and the communication time. The waiting time is the latency that the fast threads wait for the slowest thread. The communication time is the amount of time spent on synchronization. 

\begin{figure}
\centering
\includegraphics[width = 0.4\textwidth]{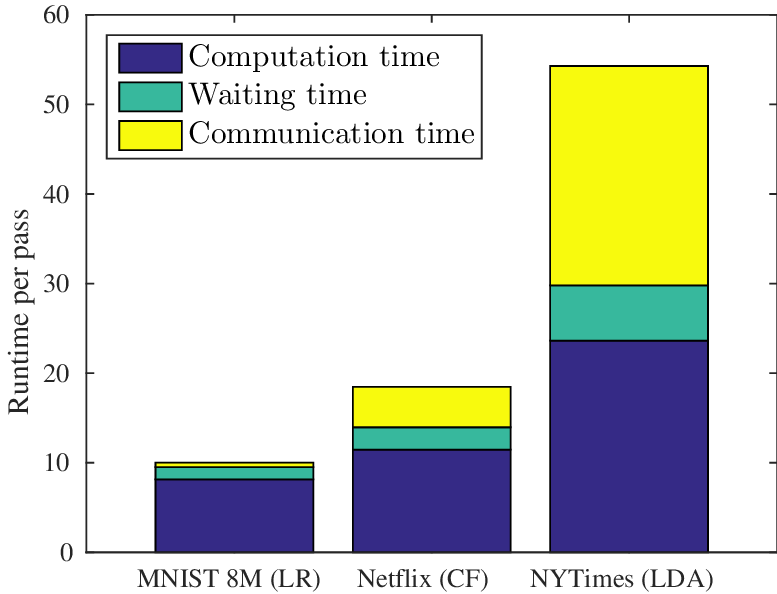}
\caption{Runtime of \splash\ for taking one pass over the training set. Three machine learning tasks: logistic regression (LR), collaborative filtering (CF) and topic modeling (LDA)}\label{fig:timeratio}
\end{figure}

We present runtime analysis on the three machine lear.ning tasks. For logistic regression and collaborative filtering, we let \splash\ workers synchronize five times per taking one pass over the training set. For topic modeling, we let the workers synchronize once per taking one pass. Figure~\ref{fig:timeratio} breaks down the runtime of \splash\ into the three parts. For the three tasks, the waiting time is 16\%, 21\% and 26\% of the computation time. This ratio will increase if the algorithm is parallelized on more machines. In contrast, the communication time is 6\%, 39\% and 103\% of the computation time --- it is proportional to the number of parameters to be learned and logarithmically proportional to the number of workers (via TreeReduce). The communication time can be reduced by decreasing the synchronization frequency.

\section{Related Work}

Distributed machine learning systems have been implemented for a variety of applications and are based on different programming paradigms. Related systems include parameter servers~\cite{power2010piccolo,dean2012large,Ahmed12scalableinference,li2014scaling}, Petuum~\cite{xing2013petuum}, Naiad~\cite{murray2013naiad} and GraphLab~\cite{low2012distributed}. There are also machine learning systems built on existing platforms, including Mahout~\cite{apache12mahout} based on Hadoop~\cite{apache12hadoop} and MLI~\cite{sparks2013mli} based on Spark~\cite{zaharia2012resilient}.

To the best of our knowledge, none of these systems are explicitly designed for parallelizing stochastic algorithms. Mahout and MLI, both adopting the iterative MapReduce~\cite{dean2008mapreduce} framework, are designed for batch algorithms. The parameter servers, Petuum and Naiad provide user-definable update primitives such as $\tt (get, set)$ on variables or $\tt (pull, push)$ on messages, under which a distributed stochastic algorithm can be implemented. However, a typical stochastic algorithm updates its parameters in every iteration, which involves expensive inter-node communication. In practice, we found that the per-iteration computation usually takes a few microseconds, but pushing an update from one Amazon EC2 node to another takes milliseconds. Thus, the communication cost dominates the computation cost. If the communication is asynchronous, 
then the algorithm will easily diverge because of the significant latency. 

GraphLab asynchronously schedules communication using a graph abstraction, which guarantees the serializability of the algorithm. Many stochastic algorithms can be written as a graph-based program. For the SGD algorithm, one constructs a vertex for every sample and every feature, and connects an edge between the vertices if the sample processes a particular feature. However, when the individual feature is shared among many samples, running SGD on this graph will cause many conflicts, which significantly restricts the degree of parallelism. Such a paradigm is efficient only if the feature is very sparse. 

\splash\ doesn't guarantee a bounded delay model like Petuum, nor does it guarantee the serializability like GraphLab. Instead, it provides a programming interface with limited operations (add, multiply and delayed add) and asks the algorithm to process weighted data. As a consequence, \splash\ exhibits descent performance in parallelizing stochastic algorithms.

Apart from distributed systems literature, there is a flurry of research studying communication-efficient methods for convex optimization. Some of them applies to stochastic algorithms. Zinkevich et al.~\cite{zinkevich2010parallelized} study the one-shot averaging scheme for parallelizing SGD. Jaggi et al.~\cite{jaggi2014communication} present a framework for parallelizing stochastic dual coordinate methods. Both methods can be implemented on top of \splash. Our theoretical analysis on SGD generalizes that of Zinkevich et al.~\cite{zinkevich2010parallelized}. In particular, the results of Section~\ref{sec:convergence} assume that the parallelized SGD is synchronized for multiple rounds, while Zinkevich et al.~\cite{zinkevich2010parallelized} let the algorithm synchronize only at the end. The multi-round synchronization scheme is more robust when the objective function is not strongly convex. But its theoretical analysis is challenging, because the updates on separate threads are no longer independent.

\section{Conclusion}

In this paper, we have described \splash, a general framework for parallelizing stochastic algorithms. The programming paradigm of \splash\ is designed around a key concept: implementing incremental updates that  processes weighted data. This paradigm allows the system to automatically parallelize the algorithm on commodity clusters. On machine learning tasks, \splash\ is order-of-magnitude faster than state-of-the-art implementations adopting the iterative MapReduce. The fast performance is partially due to the superiority of stochastic algorithms over the batch algorithm, and partially due to the communication-efficient feature of the system. In addition, \splash\ is built on top of Spark which allows it seamlessly integrating with the existing data analytics stack. 

\appendix

\section*{Appendix}
\section{Constructing Linear Transformation on a Thread}
\label{sec:construct-group-level-transformation}

When element-wise operators are sequentially applied, they merge into a single linear transformation. Assume that after processing a local subset $S$, the resulting transformation can be represented by
\[
	v \leftarrow \Gamma(S) \cdot v + \Delta(S)  + T(S)
\] 
where $\Gamma(S)$ is the scale factor,  $\Delta(S)$ is the term resulting from the element-wise add operators, and $T(S)$ is the term resulting from the element-wise delayed add operators declared before the last synchronization. 

We construct $\Gamma(S)$, $\Delta(S)$ and $T(S)$ incrementally. Let $P$ be the set of processed elements. At the beginning, the set of processed elements is empty, so that we initialize them by
\[
	\Gamma(P) = 1, \quad \Delta(P) = 0
 \quad \mbox{and} \quad \xi(P) = 0	
	\quad \mbox{for $P = \emptyset$}.
\]
After processing element $z$, we assume that the user has performed all types of operations, resulting in a transformation taking the form
\begin{align}\label{eqn:single-element-update}
	v \leftarrow \gamma (v+t) + \delta
\end{align}
where the scalars $\gamma$ and $\delta$ result from instant operators and $t$ results from the delayed operator. Concatenating transformation~\eqref{eqn:single-element-update} with the transformation constructed on set $P$, we have
\begin{align*}
	v &\leftarrow \gamma \cdot \Big( \Gamma(P)\cdot v + \Delta(P) + T(P)  + t \Big) + \delta\\
	&= \gamma \cdot \Gamma(P) \cdot v + \Big( \gamma \cdot \Delta(P) + \delta\Big) + \Big(\gamma\cdot T(P) + \gamma t\Big).
\end{align*}
Accordingly, we update the terms $\Gamma$, $\Delta$ and $T$ by
\begin{align}\label{eqn:gamma-delta-update}
	\Gamma(P\cup \{z\}) = \gamma \cdot \Gamma(P),~~ \Delta(P\cup \{z\}) = \gamma \cdot \Delta(P) + \delta~~\mbox{and}~~T(P\cup \{z\}) = \gamma \cdot T(P)+ \gamma t
\end{align}
and update the set of processed elements by $P\leftarrow P\cup\{z\}$. After processing the entire local subset, the set $P$ will be equal to $S$, so that we obtain $\Gamma(S)$, $\Delta(S)$ and $T(S)$.

\section{Determining Thread Number}
\label{sec:determine-weight}

Suppose that there are $M$ available cores in the cluster. The execution engine partitions these cores into 
several groups. Suppose that the $i$-th group contains $m_i$ cores. The group sizes are determined by the following allocation scheme:
\begin{itemize}
\item Let $4m_0$ be the thread number adopted by the last iteration. Let $4m_0 \defeq 1$ at the first iteration.
\item For $i = 1,2,\dots$, if $8m_{i-1} \leq M - \sum_{j=1}^{i-1} m_j$, the let $m_i \defeq 4m_{i-1}$. Otherwise, let $m_i\defeq M - \sum_{j=1}^{i-1} m_j$. Terminate when $\sum_{j=1}^i m_j = M$.
\end{itemize}
It can be easily verified that the candidate thread numbers (which are the group sizes) in the current iteration are at least as large as that of the last iteration. The candidate thread numbers are $4m_0,16m_0,\dots$ until they consume all of the available cores.

The $i$-th group is randomly allocated with $m_i$ Parametrized RDD partitions for training, and allocated with another $m_i$ Parametrized RDD partitions for testing. In the training phase, they execute the algorithm on $m_i$ parallel threads, following the parallelization strategy described in Section~\ref{sec:strategy}. In the testing phase, the training results are broadcast to all the partitions. The thread number associated with the smallest testing loss will be chosen. The user is asked to provide an evaluation function $\ell: W\times S\to \R$ which maps a variable-sample pair to a loss value. This function, for example, can be chosen as the element-wise loss for optimization problems, or the negative log-likelihood of probabilistic models. If the user doesn't specify an evaluation function, then the largest $m_i$ will be chosen by the system.

Once a thread number is chosen, its training result will be applied to all Parametrized RDD partitions. The allocation scheme ensures that the largest thread number is at least $3/4$ of~$M$. Thus, in case that $M$ is the best degree of parallelism, the computation power will not be badly wasted. The allocation scheme also ensures that $M$ will be the only candidate of parallelism if the last iteration's thread number is greater than $M/2$. Thus, the degree of parallelism will quickly converge to $M$ if it outperforms other degrees. Finally, the thread number is not updated in every iteration. If the same thread number has been chosen by multiple consecutive tests, then the system will continue using it for a long time, until some retesting criterion is satisfied. 

%%%%%%%%%%%%%%%%%%%%%%%%%%%%%%%%%%%%%%%%%%%%%%%%%%%%%%%%%%%%%%%%%

\section{Proof of Theorem~\ref{theorem:global-convergence}}
\label{sec:proof-global-convergence}

We assume that $\ltwos{\nabla \ell(w;x)} \leq G$ for any $(w,x)\in W\times S$. The theorem will be established if the following inequality holds:
\begin{align}\label{eqn:general-bound-recursion}
	\sum_{k=1}^T \sum_{i=1}^m \sum_{j=1}^n 2\eta_{i,j}^k \E[ L(w_{i,j}^k) - L(\wstar)]
	\leq m\E[\ltwos{w_0 - \wstar} - \ltwos{w^T - \wstar}] + G^2\sum_{k=1}^T \sum_{i=1}^m \sum_{j=1}^n (\eta_{i,j}^k)^2
\end{align}
To see how inequality~\eqref{eqn:general-bound-recursion} proves the theorem, notice that the convexity of function $L$ yields
\[
	\E[L(\wbar_j) - L(w^*)] \leq \frac{\sum_{k=1}^T \sum_{i=1}^m \sum_{j=1}^n \eta_{i,j}^k \E[ L(w_{i,j}^k) - L(\wstar)]}{\sum_{k=1}^T \sum_{i=1}^m \sum_{j=1}^n \eta_{i,j}^k}.
\]
Thus, inequality~\eqref{eqn:general-bound-recursion} implies
\[
	\E[L(\wbar_j) - L(w^*)] \leq \frac{m\E[\ltwos{w_0 - \wstar} - \ltwos{w^T - \wstar}] + G^2\sum_{k=1}^T \sum_{i=1}^m \sum_{j=1}^n (\eta_{i,j}^k)^2}{2 \sum_{k=1}^T \sum_{i=1}^m \sum_{j=1}^n \eta_{i,j}^k}.
\]
By the assumptions on $\eta_t$, it is easy to see that the numerator of right-hand side is bounded, but the denominator is unbounded. Thus, the fraction converges to zero as $T\to\infty$.

It remains to prove inequality~\eqref{eqn:general-bound-recursion}. We prove it by induction. The inequality trivially holds for $T = 0$. For any integer $k > 0$, we assume that the inequality holds for $T = k-1$. At iteration $k$, every thread starts from the shared vector $w^{k-1}$, so that $w_{i,1}^{k} \equiv w^{k-1}$. For any $j\in\{1,\dots,n\}$, let $g_{i,j}^k$ be a shorthand for $\nabla\ell(w_{i,j}^{k};x)$. A bit of algebraric transformation yields:
\begin{align*}
	\ltwos{w_{i,j+1}^{k} - \wstar}^2 &= \ltwos{\Pi_W(w_{i,j}^{k} - \eta_{i,j}^k g_{i,j}^k)- \wstar}^2 \leq \ltwos{w_{i,j}^{k} - \eta_{i,j}^k g_{i,j}^k - \wstar}^2\\
	&= \ltwos{w_{i,j}^{k} - \wstar}^2 + (\eta_{i,j}^k)^2\ltwos{g_{i,j}^k}^2
	- 2\eta_{i,j}^k \langle w_{i,j}^{k} - \wstar, g_{i,j}^k\rangle,
\end{align*}
where the inequality holds since $\wstar\in W$ and $\Pi_W$ is the projection onto $W$. Taking expectation on both sides of the inequality and using the assumption
that $\ltwos{g_{i,j}^k} \leq G$, we have
\begin{align*}
	\E[\ltwos{w_{i,j+1}^{k} - \wstar}^2] \leq \E[\ltwos{w_{i,j}^{k} - \wstar}^2]
	+ G^2(\eta_{i,j}^k)^2 - 2\eta_{i,j}^k \E[\langle w_{i,j}^{k} - \wstar, \nabla L(w_{i,j}^{k})\rangle].
\end{align*}
By the convexity of function $L$, we have $\langle w_{i,j}^{k} - \wstar, \nabla L(w_{i,j}^{k})\rangle \geq L(w_{i,j}^{k}) - L(\wstar)$. Plugging in this inequality, we have
\begin{align}\label{eqn:basic-inequality-general-sgd}
	2\eta_{i,j}^k \E[L(w_{i,j}^{k}) - L(\wstar)] \leq \E[\ltwos{w_{i,j}^{k} - \wstar}^2]
	- \E[\ltwos{w_{i,j+1}^{k} - \wstar}^2] +  G^2(\eta_{i,j}^k)^2.
\end{align}
Summing up inequality~\eqref{eqn:basic-inequality-general-sgd} for $i=1,\dots,m$
and $j = 1,\dots,n$, we obtain
\begin{align}\label{eqn:summed-inequality-general-sgd}
	\sum_{i=1}^m\sum_{j=1}^n 2\eta_{i,j}^k \E[L(w_{i,j}^{k}) - L(\wstar)] &\leq m\E[\ltwos{w^{k-1} - \wstar}^2] - \sum_{i=1}^m \E[\ltwos{w_{i,n+1}^{k} - \wstar}^2] \nonumber\\
	&\qquad +  \sum_{i=1}^m\sum_{j=1}^n G^2(\eta_{i,j}^k)^2.
\end{align}
Notice that $w^k = \frac{1}{m} w_{i,n+1}^{k}$. Thus, Jensen's inequality implies $\sum_{i=1}^m \ltwos{w_{i,n+1}^{k} - \wstar}^2 \geq m \ltwos{w^k - \wstar}$. Plugging this inequality to upper bound~\eqref{eqn:summed-inequality-general-sgd} yields
\begin{align}\label{eqn:overall-inequality-general-sgd}
	\sum_{i=1}^m\sum_{j=1}^n 2\eta_{i,j}^k \E[L(w_{i,j}^{k}) - L(\wstar)] \leq m\E[\ltwos{w^{k-1} - \wstar}^2 - \ltwos{w^k - \wstar}^2] +  \sum_{i=1}^m\sum_{j=1}^n G^2(\eta_{i,j}^k)^2.
\end{align}
The induction is complete by combining upper bound~\eqref{eqn:overall-inequality-general-sgd} with the inductive hypothesis.

\section{Proof of Theorem~\ref{theorem:strong-convex-sgd}}
\label{sec:proof-strong-convex-sgd}

Recall that $w^k$ is the value of vector $w$ after iteration $k$. Let $w_{i}^k$ be the output of thread $i$ at the end of iteration $k$. According to the update formula, we have $w^k = \
\Pi_B(\frac{1}{m}\sum_{i=1}^m w_i^k)$, where $\Pi_B(\cdot)$
is the projector to the set $B$. The set $B$ contains the optimal solution $\wstar$. Since projecting to a convex set doesn't increase the point's distance to the elements in the set, and because that $w_{i}^k$ ($i=1,\dots,m$) are mutually independent conditioning on $w^{k-1}$, we have
\begin{align}\label{eqn:decompose-bias-var}
\E[\ltwos{w^k - \wstar}^2] &\leq \E\Big[\E\Big[\Big\|\frac{1}{m}\sum_{i=1}^m w_i^k - \wstar\Big\|_2^2 \Big | w^{k-1} \Big]\Big]\nonumber\\
& = \frac{1}{m^2}\sum_{i=1}^m \E[\E[\ltwos{w_i^k - \wstar}^2 | w^{k-1}]]
+ \frac{1}{m^2} \sum_{i\neq j}\E[\E[\langle w_i^k - \wstar, w_j^k - \wstar \rangle | w^{k-1}]]\nonumber\\
&= \frac{1}{m} \E[\ltwos{w_1^k - \wstar}^2]
+ \frac{m-1}{m} \E[\ltwos{\E[w_1^k|w^{k-1}] - \wstar}^2]
\end{align}
Equation~\eqref{eqn:decompose-bias-var} implies that we could upper bound the two terms on the right-hand side respectively. To this end, we introduce three shorthand notations:
\begin{align*}
	&a_k \defeq \E[\ltwos{w^k - \wstar}^2],\\
	&b_k \defeq \E[\ltwos{w_1^k - \wstar}^2],\\
	&c_k \defeq \E[\ltwos{\E[w_1^k|w^{k-1}] - \wstar}^2].
\end{align*}
Essentially, equation~\eqref{eqn:decompose-bias-var} implies
$a_k \leq \frac{1}{m}b_k + \frac{m-1}{m}c_k$. Let $a_0\defeq \ltwos{w^0 - \wstar}$ where $w^0$ is the initial vector. 
The following two lemmas upper bounds $b_{k+1}$
 and $c_{k+1}$ as functions of $a_{k}$. We defer their proofs to the
 end of this section.
 
\begin{lemma}\label{lemma:sgd-variance}
For any integer $k\geq 0$, we have
\begin{align*}
	b_{k+1} \leq \frac{k^2}{(k+1)^2}a_k + \frac{\beta_1}{ (k+1)^2n}
	\quad\mbox{where}\quad \beta_1\defeq 4 G^2/ \lambda^2.
\end{align*}
\end{lemma}

\begin{lemma}\label{lemma:sgd-bias}
We have $c_1 \leq \beta_2^2/n^2$ and for any integer $k\geq 1$,
\begin{align*}
	c_{k+1} \leq \frac{k^2}{(k+1)^2}a_k + \frac{2\beta_2 \sqrt{a_k} + \beta_2^2/n}{(k+1)^2n}
\quad \mbox{where} \quad \beta_2 \defeq \max\left\{
	\lceil 2 H/\lambda \rceil R, \frac{8G^2(L+G/\rho^2)}{\lambda^3} \right\}.
\end{align*}
\end{lemma}

Combining equation~\eqref{eqn:decompose-bias-var} with the results of
Lemma~\eqref{lemma:sgd-variance} and Lemma~\eqref{lemma:sgd-bias}, we obtain an upper bound on $a_1$:
\begin{align}\label{eqn:multi-round-sgd-init}
	a_1 \leq \frac{\beta_1}{mn} + \frac{\beta_2^2}{n^2} \defeq \beta_3.
\end{align}
Furthermore, Lemma~\eqref{lemma:sgd-variance} and Lemma~\eqref{lemma:sgd-bias} upper bound $a_{k+1}$ as a function of $a_k$:
\begin{align}\label{eqn:multi-round-sgd-recurssion}
	a_{k+1} \leq \frac{k^2}{(k+1)^2}a_k + \frac{\beta_3 + 2\beta_2\sqrt{a_k}/n}{(k+1)^2}.
\end{align}
Using upper bounds~\eqref{eqn:multi-round-sgd-init} and~\eqref{eqn:multi-round-sgd-recurssion}, we claim that
\begin{align}\label{claim:multi-round-sgd-bound}
	a_k \leq \frac{\beta_3 + 2\beta_2\sqrt{\beta_3}/n}{k}  \quad \mbox{for}\quad k = 1,2,\dots
\end{align}
By inequality~\eqref{eqn:multi-round-sgd-init}, the claim is true for $k = 1$. We assume that the claim holds for $k$ and prove it for $k+1$.
Using the inductive hypothesis, we have $a_k \leq \beta_3$. Thus,
inequality~\eqref{eqn:multi-round-sgd-recurssion} implies
\begin{align*}
	a_{k+1} \leq \frac{k^2}{(k+1)^2}\cdot \frac{\beta_3 + 2\beta_2\sqrt{\beta_3}/n}{k} + \frac{\beta_3 + 2\beta_2\sqrt{\beta_3}/n}{(k+1)^2} = \frac{\beta_3 + 2\beta_2\sqrt{\beta_3}/n}{(k+1)n}
\end{align*}
which completes the induction. Note that both $\beta_1$ and $\beta_2$
are constants that are independent of $k$, $m$ and $n$. Plugging the definition of $\beta_3$, we can rewrite inequality~\eqref{claim:multi-round-sgd-bound}
as
\begin{align*}
	a_k \leq \frac{4G^2}{\lambda^2 k m n} + \frac{C_1}{k m^{1/2}n^{3/2}}
	+ \frac{C_2}{k n^{2}}.
\end{align*}
where $C_1$ and $C_2$ are constants that are independent of $k$, $m$ and $n$. This completes the proof of the theorem.

\subsection{Proof of Lemma~\ref{lemma:sgd-variance}}

In this proof, we use $w_j$ as a shorthand to denote the value of vector $w$ at iteration $k+1$ when the first thread is processing the $j$-th element. We drop the notation's dependence on the iteration number and on the thread index since they are explicit from the context. Let $g_j = \nabla \ell(w_j;x_j)$ be the gradient of loss function $\ell$ with respect to $w_j$ on the $j$-th element. Let $\eta_j$ be the stepsize parameter when $w_j$ is updated. It is easy to verify that $\eta_j = \frac{2}{\lambda(kn+j)}$.

We start by upper bounding the expectation of $\ltwos{w_1^{k+1} - \wstar}^2$ conditioning on $w^k$. 
By the strong convexity of $L$ and the fact that $\wstar$ minimizes $L$, we have
\begin{align*}
    \langle \E[g_j], w_j-\wstar \rangle \geq L(w_j) - L(\wstar) + \frac{\strongparam}{2}\ltwo{w_j-\wstar}^2.
\end{align*}
as well as
\begin{align*}
  L(w_j) - L(\wstar) \geq \frac{\strongparam}{2}\ltwo{w_j-\wstar}^2.
\end{align*}
Hence, we have 
\begin{align}\label{eqn:inner-grad-vecdiff}
\langle \E[g_j], w_j-\wstar \rangle \geq \strongparam\ltwo{w_j-\wstar}^2
\end{align}
Recall that $\Pi_W(\cdot)$
denotes the projection onto set $W$. By the convexity of $W$, we
have $\ltwo{\Pi_W(u)-v}\leq \ltwo{u-v}$ for any $u,v\in W$. Using these inequalities, we have
the following:
\begin{align*}
  \E[\ltwos{w_{j+1}-\wstar}^2|w^{k}] &= \E[\ltwos{\Pi_W(w_j-\eta_jg_j)-\wstar}^2|w^{k}]\\
  &\leq \E[\ltwos{w_j-\eta_jg_j -\wstar }^2|w^{k}]\\
  & = \E[\ltwos{w_j-\wstar}^2|w^{k}] - 2\eta_j\E\left[\langle g_j, w_j-\wstar \rangle|w^{k}\right] + \eta_j^2\E[\ltwos{g_j}^2|w^{k}].
\end{align*}
Note that the gradient $g_j$ is
independent of $w_j$ conditioning on $w^{k-1}$. Thus, we have
\[
\E\left[\langle g_j, w_j-\wstar \rangle|w^{k}\right]
= \E[\langle \E[g_j], w_j-\wstar  \rangle | w^{k}] \geq \strongparam \E[\ltwo{w_j-\wstar}^2 | w^{k}].
\]
where the last inequality follows from inequality~\eqref{eqn:inner-grad-vecdiff}. As a consequence, we have
\begin{align*}
\E[\ltwos{w_{j+1}-\wstar}^2|w^{k}] \leq (1-2\stepsize_j\strongparam)\E[\ltwos{w_j-\wstar}^2|w^{k}] + \stepsize_j^2\lipobj^2.
\end{align*}
Plugging in $\stepsize_j = \frac{2}{\lambda(kn + j)}$, we obtain
\begin{align}\label{eqn:simple-sgd-induction}
  \E[\ltwos{w_{j+1}-\wstar}^2|w^{k}] \leq \left(1 - \frac{4}{kn+j}\right)\E[\ltwos{w_j-\wstar}^2|w^{k}] + \frac{4\lipobj^2}{\lambda^2 (kn+j)^2}.
\end{align}

\paragraph{Case $k=0$:} We claim that any $j\geq 1$,
\begin{align}\label{claim:k-eq-1-var}
\E[\ltwos{w_{j}-\wstar}^2] \leq \frac{4G^2}{\lambda^2 j}
\end{align} 
Since $w^1_1 = w_{n+1}$, the claim establishes the lemma.
We prove the claim by induction. The claim holds for $j=1$ because inequality~\eqref{eqn:inner-grad-vecdiff}
yields
\[
	\ltwos{w_{1}-\wstar}^2 \leq \frac{\langle\E[g_1],w_1-\wstar\rangle}{\lambda}
	\leq \frac{G\ltwos{w_{1}-\wstar}}{\lambda} \quad
	\Rightarrow \quad \ltwos{w_{1}-\wstar} \leq G/\lambda.
\]
Otherwise, we assume that the claim holds for $j$. Then inequality~\eqref{eqn:simple-sgd-induction} yields
\begin{align*}
	\E[\ltwos{w_{j+1}-\wstar}^2] &\leq \left(1 - \frac{4}{j}\right)
	\frac{4 G^2}{\lambda^2 j} + \frac{4 G^2}{\lambda^2 j^2}\\
	&= \frac{4 G^2}{\lambda^2}\frac{j-4+1}{j^2} \leq \frac{4 G^2}{\lambda^2 (j+1)},
\end{align*}
which completes the induction.

\paragraph{Case $k>0$:} We claim that for any $j\geq 1$,
\begin{align}\label{claim:k-gtr-1-var}
\E[\ltwos{w_{j}-\wstar}^2|w^k] \leq \frac{1}{(kn+j-1)^2}\left( (kn)^2\ltwos{w^{k}-\wstar}^2 + \frac{4 G^2(j-1)}{\lambda^2}\right)
\end{align}
We prove~\eqref{claim:k-gtr-1-var} by induction. The claim is obviously true for $j=1$. Otherwise, we assume that the claim holds for $j$  and prove it for $j+1$. Since
$1-\frac{4}{kn+j} \leq (\frac{kn+j-1}{kn+j})^2$, combining the inductive hypothesis and inequality~\eqref{eqn:simple-sgd-induction}, we have
\begin{align*}
 &\E[\ltwos{w_{j+1}-\wstar}^2 | w^{k}]\\
 &\qquad \leq \frac{1}{(kn+j)^2}\left( (kn)^2\ltwos{w^{k}-\wstar}^2 + \frac{4G^2(j-1)}{\lambda^2}\right) + \frac{4 \lipobj^2}{\lambda^2 (kn+j)^2}\\
 &\qquad = \frac{1}{(kn+j)^2}\left( (kn)^2\ltwos{w^{k}-\wstar}^2 + \frac{4 G^2j}{\lambda^2}\right).
\end{align*}
which completes the induction. Note that claim~\eqref{claim:k-gtr-1-var} establishes the lemma since $w^k_1 = w_{n+1}$.

%%%%%%%%%%%%%%%%%%%%%%%%%%%%%%%%%%%%%%%%%%%%%%%%%%%%%%%%%%%%%%
\subsection{Proof of Lemma~\ref{lemma:sgd-bias}}
In this proof, we use $w_j$ as a shorthand to denote the value of vector $w$ at iteration $k+1$ when the first thread is processing the $j$-th element. We drop the notation's dependence on the iteration number and on the thread index since they are explicit from the context. Let $g_j = \nabla \ell(w_j;x_j)$ be the gradient of loss function $\ell$ with respect to $w_j$ on the $j$-th element. Let $\eta_j$ be the stepsize parameter when $w_j$ is updated. It is easy to verify that $\eta_j = \frac{2}{\lambda(kn+j)}$.

Recall the
neighborhood $U_\rho \subset W$ in
Assumption~\ref{assumption:parameter-space}, and note that
\begin{align*}
  w_{j+1} - \wstar
  & = \Pi_W(w_j - \eta_j g_j - \wstar) \\
  & = w_j - \eta_j g_j - \wstar
  + \indicator(w_{j+1} \not\in U_\rho)
  \left(\Pi_W(w_j - \eta_j g_j)
  - (w_j - \eta_j g_j)\right)
\end{align*}
since when $w \in  U_\rho$, we have $\Pi_W(w) = w$.
Consequently, an application of the triangle inequality and Jensen's inequality gives
\begin{align*}
  \ltwos{\E[w_{j+1}-\wstar|w^k]}
  &\leq \ltwos{\E[w_j-\eta_jg_j-\wstar|w^k]}\\
  &\quad + \E\left[\ltwo{(\Pi_W(w_j-\eta_jg_j)
      -(w_j-\eta_jg_j))1(w_{j+1}\notin U_\rho)}|w^k\right].
\end{align*}
By the definition of the projection and the fact that $w_j \in
W$, we additionally have
\begin{equation*}
  \ltwo{\Pi_W(w_j-\eta_jg_j)
    -(w_j-\eta_jg_j)}
  \le \ltwo{w_j-(w_j-\eta_jg_j))}
  \le \eta_j \ltwo{g_j}.
\end{equation*}
Thus, by combining the above two inequalities, and applying Assumption~\ref{assumption:smoothness-sgd}, we have
\begin{align}
  \ltwos{\E[w_{j+1}-\wstar|w^k]}
  & \le \ltwos{\E[w_j-\eta_jg_j-\wstar|w^k]}
  + \eta_j \E[\ltwo{g_j} \indic{w_{j+1} \not\in U_\rho}|w^k]
  \nonumber \\
  & \le \ltwos{\E[w_j-\eta_jg_j-\wstar|w^k]}
  + \eta_j \lipobj\cdot
  P(w_j \not\in U_\rho|w^k)
  \nonumber \\
  & \le \ltwos{\E[w_j-\eta_jg_j-\wstar|w^k]}
  + \eta_j \lipobj\cdot \frac{\E[\ltwos{w_{j+1}-\wstar}^2|w^k]}{\rho^2},
  \label{eqn:sgd-tplusone-truncation}
\end{align}
where the last inequality follows from the Markov's inequality.

Now we turn to controlling the rate at which $w_j - \eta_j -g_j
$ goes to zero. Let $\ell_j(\cdot) = \ell(\cdot; x_j)$ be a
shorthand for the loss evaluated on the $j$-th data element. By
defining
\begin{align*}
r_j \defeq g_j - \nabla \ell_j(\wstar) - \nabla^2
  \ell_j(\wstar)(w_j - \wstar),
\end{align*}
a bit of algebra yields
\begin{equation*}
  g_j = \nabla \ell_j(\wstar)
  + \nabla^2 \ell_j(\wstar)(w_j-\wstar) + r_j.
\end{equation*}
First, we note that $\E[\nabla \ell_j(\wstar)|w^k] = \nabla L(\wstar) = 0$. Second, the Hessian $\nabla^2 \ell_j(\wstar)$ is
independent of $w_j$. Hence we have
\begin{align}
  \E[g_j|w^k] &= \E[\nabla \ell_j(\wstar)] + \E[\nabla^2 \ell_j(\wstar)|w^k]\cdot\E[w_j-\wstar|w^k]+ \E[r_j|w^k]\nonumber\\
  &= \nabla^2 L(\wstar)\E[w_j-\wstar|w^k]+ \E[r_j|w^k].
  \label{eqn:sgd-truncate-grad}
\end{align}
Taylor's theorem implies that $r_j$ is the
Lagrange remainder
\begin{align*}
  r_j = (\nabla^2 \ell_j(w') - \nabla^2 \ell_j(\wstar))(w'-\wstar),
\end{align*}
where $w'= \interp w_j +(1-\interp)\wstar$ for
some $\interp \in [0, 1]$.
Applying Assumption~\ref{assumption:smoothness-sgd}, we find that
\begin{align}
  \E[\ltwos{r_j}|w^k]
  & \le \E[
    \ltwos{\nabla^2 \ell_j(w') - \nabla^2 \ell_j(\wstar)}
    \ltwo{w_j - \wstar}|w^k]\nonumber \\
  & \le \liphessian \E[\ltwos{w_j - \wstar}^2|w^k].\label{eqn:sgd-rt-bound}
\end{align}
By combining the expansion~\eqref{eqn:sgd-truncate-grad} with the
bound~\eqref{eqn:sgd-rt-bound}, we find that
\begin{align*}
  \lefteqn{\ltwos{\E[w_j - \eta_j g_j - \wstar|w^k]}
    = \ltwo{\E[(I - \eta_j \nabla^2 F_0(\wstar))(w_j - \wstar)
        + \eta_j r_j|w^k]}} \\
  & \qquad \qquad
  \le \ltwos{(I - \eta_j \nabla^2 L(\wstar))\E[w_j - \wstar|w^k]}
  + \eta_j \liphessian \E[\ltwos{w_j - \wstar}^2|w^k].
\end{align*}
Using the earlier bound~\eqref{eqn:sgd-tplusone-truncation} and plugging in the assignment $\eta_j = \frac{2}{\lambda(kn+j)}$, this inequality
then yields
\begin{align}
  \ltwos{\E[w_{j+1} - \wstar|w^k]}
  &\le \ltwo{I - \eta_j \nabla^2 L(\wstar)}
  \ltwos{\E[w_j - \wstar|w^k]}\nonumber\\
  &\quad + \frac{2}{\strongparam (kn+j)}\left(L\E[\ltwos{w_j - \wstar}^2|w^k] + \frac{G\E[\ltwos{w_{j+1}-\wstar}^2|w^k]}{\rho^2} \right).\label{eqn:wjp1-minus-wstar}
\end{align}
Next, we split the proof into two cases when $k=1$ and $k > 1$.

\paragraph{Case $k=0$:} Note that by strong convexity and
our condition that $\ltwos{\nabla^2 L(\wstar)} \le \lipgrad$, whenever $\eta_j \lipgrad \leq 1$ we
have
\begin{equation*}
  \ltwos{I - \eta_j \nabla^2 L(\wstar)}
  = 1 - \eta_j \lambda_{\min}(\nabla^2 L(\wstar))
  \le 1 - \eta_j \strongparam
\end{equation*}
Define $\tau_0 = \ceil{2 \lipgrad /\strongparam}$; then for $j \ge \tau_0$, we have $ \eta_j \lipgrad\leq 1$. As a consequence, inequality~\eqref{claim:k-eq-1-var} (in the proof of Lemma~\ref{lemma:sgd-variance})
and inequality~\eqref{eqn:wjp1-minus-wstar} yield that for any $j \ge \tau_0$,
\begin{equation}
  \label{eqn:sgd-induction}
  \ltwo{\E[w_{j+1}-\wstar]}
  \le (1-2/j)\ltwo{\E[w_j-\wstar]}
  + \frac{8 \lipobj^2}{\strongparam^3 j^2}\left(L + G/\rho^2\right).
\end{equation}
As shorthand notations, we define two intermediate variables
\begin{align*}
  u_{t}= \ltwo{\E(w_j-\wstar)}
  ~~~\mbox{and}~~~
  b_1 = \frac{8 \lipobj^2}{\strongparam^3}\left(L + G/\rho^2\right).
\end{align*}
Inequality \eqref{eqn:sgd-induction} then implies the inductive
relation
\[
    u_{j+1}\leq (1-2/j)u_j + b_1/j^2 \quad \mbox{for any $j\geq \tau_0$.}
\]
Now we claim that by defining $b_2 \defeq \max\{\tau_0 \radius, b_1\}$, we have $u_j \le \beta / j$. Indeed, it is clear that $u_j \le \tau_0 \radius/j$ for $j=1,2,\dots,\tau_0$.
For $t > \tau_0$, using the inductive hypothesis, we have
\begin{align*}
  u_{j+1} &\le
  \frac{(1 - 2/j) b_2}{j}
  + \frac{b_1}{j^2}
  \leq \frac{b_2 j - 2 b_2 + b_2}{j^2}= \frac{b_2(j - 1)}{j^2}
  \le \frac{b_2}{j + 1}.
\end{align*}
This completes the induction and establishes the lemma for $k=0$.

\paragraph{Case $k>0$:} Let $u_j= \ltwos{\E[w_j-\wstar | w^k]}$
and $\delta = \ltwos{w^k-\wstar}$ as shorthands. Combining inequality~\eqref{claim:k-gtr-1-var} (in the proof of Lemma~\ref{lemma:sgd-variance})
and inequality~\eqref{eqn:wjp1-minus-wstar} yield
\begin{align}
	u_{j+1} &\leq \left( 1 - \frac{2}{kn+j}\right) u_j + \frac{2(L + G/\rho^2)}{\strongparam (kn+j)(kn+j-1)^2}\left( (kn)^2\delta^2 + \frac{4G^2j}{\lambda^2}\right)\nonumber\\
	&\leq \left( 1 - \frac{2}{kn+j}\right) u_j + \frac{2(L + G/\rho^2)}{\strongparam (kn+j)(kn+j-1)kn}\left( (kn)^2\delta^2 + \frac{4 G^2n}{\lambda^2}\right)\nonumber\\
	&= \frac{(kn+j-2)(kn+j-1)}{(kn+j-1)(kn+j)} u_j + \frac{b_1 kn \delta^2 + b_2/k}{(kn+j-1)(kn+j)}\label{eqn:k-gtr-1-bias-recursion}
\end{align}
where we have introduced shorthand notations $b_1 \defeq \frac{2(L+G/\rho^2)}{\lambda}$ and $b_2 \defeq \frac{8 G^2(L+G/\rho^2)}{\lambda^3}$. With these notations, we claim that
\begin{align}
u_j \leq \frac{(kn-1)kn \delta + (j-1)(b_1 kn \delta^2 + b_2/k)}{(kn+j-2)(kn+j-1)}.
\end{align}
We prove the claim by induction. Indeed, since $u_1 = \delta$, the claim obviously holds for
$j = 1$. Otherwise, we assume that the claim holds for $j$, then
inequality~\eqref{eqn:k-gtr-1-bias-recursion} yields
\begin{align*}
u_{j+1} &\leq \frac{(kn-1)kn \delta + (j-1)(b_1 kn \delta^2 + b_2/k)}{(kn+j-1)(kn+j)} + \frac{b_1 kn \delta^2 + b_2/k}{(kn+j-1)(kn+j)}\\
&=\frac{(kn-1)kn \delta + j(b_1 kn \delta^2 + b_2/k)}{(kn+j-1)(kn+j)},
\end{align*}
which completes the induction. As a consequence, a bit of algebraic transformation yields
\begin{align}
\ltwos{\E[w^{k+1}_1-\wstar|w^k]} &= u_{n+1} \leq \frac{(kn-1)kn \delta + n(b_1 kn \delta^2 + b_2/k)}{((k+1)n-1)(k+1)n}\nonumber\\
&\leq \frac{k^2n^2\delta}{(k+1)^2n^2} 
+ \frac{nb_1kn\delta^2}{kn(k+1)n} 
+ \frac{nb_2/k}{kn(k+1)n} 
\nonumber\\
&\leq \left(\frac{k}{k+1}\right)^2\delta + \frac{b_1\delta^2}{k+1}
+ \frac{b_2}{k(k+1)n}\nonumber\\
&= \frac{k}{k+1}\left( \frac{k\delta + \frac{k+1}{k} b_1\delta^2}{k+1} + \frac{b_2}{k^2n}\right)\label{eqn:wkp11-minus-wstar}
\end{align}
By the fact that $w^k\in B$, we have $\frac{k+1}{k} b_1\delta\leq \frac{k+1}{k} b_1 D \leq 1$.
Thus, inequality~\eqref{eqn:wkp11-minus-wstar} implies
\begin{align*}
\ltwos{\E[w^{k+1}_1-\wstar|w^k]}^2 \leq \left(\frac{k}{k+1}\right)^2\left(\delta + \frac{b_2}{k^2n}\right)^2
\end{align*}
Taking expectation on both sides of the inequality, then applying Jensen's inequality, we obtain
\begin{align*}
\E[\ltwos{\E[w^{k+1}_1-\wstar|w^k]}^2] \leq \frac{k^2\E[\delta^2]}{(k+1)^2} + \frac{2b_2 \sqrt{\E[\delta^2]} + b_2^2/ n}{(k+1)^2 n}.
\end{align*}
Hence, the lemma is established.

\bibliographystyle{abbrv} \bibliography{bib}
\end{document}